\documentclass[conference]{IEEEtran}
\IEEEoverridecommandlockouts
\usepackage{subfigure} 
\usepackage{cite}
\usepackage{balance}
\usepackage{amsmath,amssymb,amsfonts}
\usepackage{algorithmic}
\usepackage{graphicx}
\usepackage{textcomp}
\usepackage{xcolor}
\usepackage[utf8x]{inputenc}
\usepackage{mathrsfs}
\usepackage{tabularx}  
\usepackage{bm}
\usepackage{color}

\def\BibTeX{{\rm B\kern-.05em{\sc i\kern-.025em b}\kern-.08em
    T\kern-.1667em\lower.7ex\hbox{E}\kern-.125emX}}

\begin{document}

\title{Road Curb Detection Using A Novel Tensor Voting Algorithm}
\author{Yilong Zhu$^{1}$,Dong Han$^{2}$,Bohuan Xue$^{1}$,Jianhao Jiao$^{1}$,Zuhao Zou$^{2}$,Ming Liu$^{1}$, Rui Fan$^{1}$\\
$^{1}$Robotics and Multi-Perception Laborotary, Robotics Institute,\\  Hong Kong University of Science and Technology, Clear Water Bay, Hong Kong SAR, China.\\
$^{2}$Shenzhen Institutes of Advanced Technology, Chinese Academy of Sciences\\
Email:  yzhubr@connect.ust.hk, dong.han@siat.ac.cn, \{bxueaa, hhuangat, eelium, eeruifan\}@ust.hk
\vspace{-1.0em}
}

\maketitle

\begin{abstract}

Road curb detection is very important and neces-sary for autonomous driving because it can improve the safety and robustness of robot navigation in the outdoor environment. In this paper, a novel road curb detection method based on tensor voting is presented. The proposed method processes the dense point cloud acquired using a 3D LiDAR. Firstly, we utilize a sparse tensor voting approach to extract the line and surface features. Then, we use an adaptive height threshold and a surface vector to extract the point clouds of the road curbs. Finally, we utilize the height threshold to segment different obstacles from the occupancy grid map. This also provides an effective way of generating high-definition maps. The experimental results illustrate that our proposed algorithm can detect road curbs with near real-time performance.  
\end{abstract}

\section{Introduction}
\label{sec.introduction}
\subsection{Motivation}
Road curbs are one of the most common road features. The accurate detection of road curb can provide a traversable area for unmanned ground vehicle (UGV) \cite{kim2007traversability}. Furthermore, when we plan a path for a UGV, the road curbs have to be detected beforehand because this can help the UGV avoid collision \cite{zhao2011dynamic}. Furthermore, the information of the road curb is usually utilized in adaptive Monte Carlo localization (AMCL) \cite{thrun2001robust} to provide a constraint in the axis is perpendicular to the road curb.

Currently, most researchers geometry to extract road curbs. The method proposed in this paper is developed based on temporal filters and spline fitting.

However, most of the algorithms require some expen-sive Lidar, such as, Velodyne HDL-64-E, StreetMapper, and RIEGL VMX-250. These algorithms are usually very computationally intensive \cite{puente2013review}.

Halawany proposed algorithm \cite{el2011detection} computes the surface normal and the normalized eigenvalues to extract features, such as road curbs. The output of our curb detection system is also utilized as the input of AMCL for localization.
\subsection{Contributions}
We address two issues in this paper:
 \begin{itemize}
 \item{}
  We present a novel method to extract the point clouds of road curbs from dense point clouds generated using the tensor voting frame-work. Firstly, we extract the line and surface features using tensor voting and use the surface feature to estimate the ground height, use the normal information and the line feature to extract the curb.
  \item{}After getting the information of road curbs, we also use the digital elevation map (DEM) to estimate different obstacles with different traversable features. We also use different colors to represent the different semantically meaningful regions in the grid map for further path planning.
 \end{itemize}
\subsection{Organization}
The remainder of this paper is organized as follows. We present related works in Section II. In Section III, we briefly introduce our system. In Section IV, we compare two different ways of obtaining the dense cloud map: iterative closest point (ICP) mapping and lidar odometry and mapping (LOAM). The Section V gives the method to detect the road curb by tensor voting. In VI; we use the result of tensor voting to extract the road curb and project it to a grid map with different semantics.

\section{Related Works}
\subsection{Road Curb Detection}
The approaches for road curb detection can be classified into two main categories: geometry-based and machine-lerning-based. Zhao et al. \cite{zhao2012curb} use three spatial cues and a parabola model to detect curbs. Hata et al. \cite{hata2014robust} introduce a least trimmed squares (LTS)-based curb detection approach. In addition, Kodagoda et al. \cite{kodagoda2002road} presents an effective detection method based on an extended Kalman filter (EKF) to esti-mate the positions of road curbs. Kellner et al. \cite{kellner2014road} presents a method based on a digital elevation map (DEM) and this method estimates the most likely path to get the curb. Fong et al. \cite{fong2003representing} also utilize DEM for the same purpose. Furthermore, Caltagirone et al. \cite{caltagirone2017fast} introduces a robust algorithm based on a convolutional neural network (CNN) to detect road surface and traversable areas from bird’s-eye view maps. This shows good performance in terms of both speed and accuracy.
\subsection{Dense Point Cloud Map}
There are lots of ways to get the direct dense point cloud map. The first way is to get the dense point cloud map from the sensor in one frame; for example, we can get the dense cloud map from the stereo camera and reconstruct the road surface \cite{fan2018road} \cite{fan2018real}, and we also can get the dense point cloud map from solid-state Lidar.

The other way to get a dense point cloud map is to get the point from several frames, one example is LiDAR SLAM, where we can get the dense point cloud map by the LOAM algorithm given by Zhang \cite{zhang2014loam}, or an algorithm like ICP \cite{pomerleau2013comparing} mapping.
\subsection{Tensor Voting}
The tensor voting \cite{medioni2000computational} starts with computer vision; and it is widely used in feature extraction; it shows a good tolerance in noise. Schuster \cite{schuster2004segmentation} uses this method for point cloud surface segmentation. In our attempt; we use the GPU to accelerate the computing process according to the method given in \cite{liu2012normal}. This method can extract the different features in the image and point clouds.
\subsection{Digital Elevation Map}
 The digital elevation map is widely used in ground finding and road curb detection. Fong \cite{fong2003representing} uses a digital elevation map to get the traversable area and make a two and one-half dimensional map structure to store the height of each grid’s height. Kellner \cite{kellner2014road} gives a new way to obtain the digital elevation map with a motion estimation module. Some scholars have attempted using a Kalman-Filter to estimate the cells height value \cite{cremean2006system}. It is important to know the road height for the UGV in an urban environment, for which the classification and tracking can be done based on the road height.

\section{SYSTEM ARCHITECTURE}

\subsection{Sensors and platforms}
\begin{figure}[t!]
	\centering
	\includegraphics[width=0.40\textwidth]{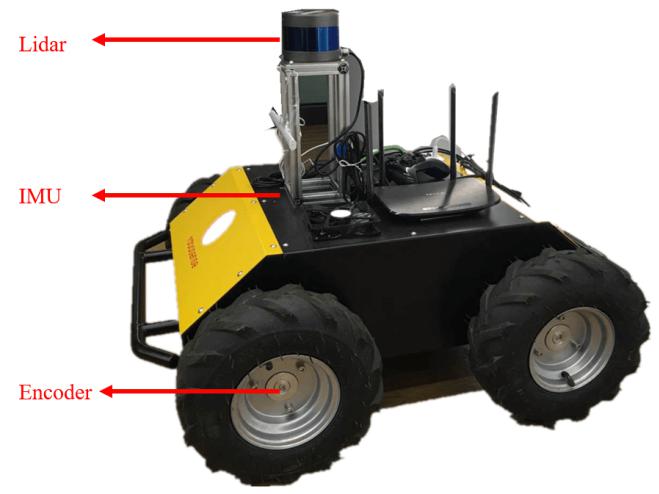}
	\caption{An example of the UGV platform.}
	\vspace{-1.5em}
\end{figure}

\begin{figure}[t!]
	\centering
	\includegraphics[width=0.5\textwidth]{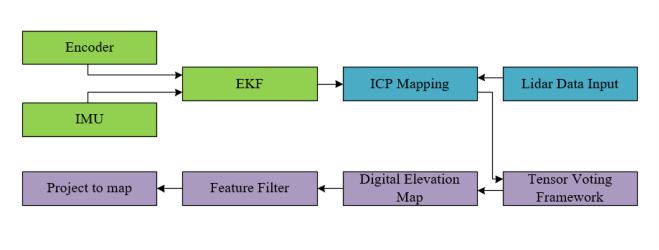}
	\caption{Block diagram of the proposed system.}
	\vspace{-0.5em}
\end{figure}

Our method is evaluated on a UGV. This platform is a differential drive ground robot. An example of the platform is shown in Fig. 1.

The sensors and perception modules we use include Li-DAR, an inertial measurement unit (IMU), and an encoder. In addition, the vehicle is equipped with an RS-LiDAR-16 produced by Robosense which can update the data with 20 Hz.

\subsection{Program architecture}

In this paper, our program has 4 major parts, which are shown in Fig. 2. The first part is the mapping part, in which fuse the encoder and IMU to get the pose estimation of our UGV by the EKF. The encoder and IMU are light-invariant sensors, so we can get the exact odometry both day and night. The Lidar works at 20 Hz, and during the gap we should use the result of the EKF to get the shift of the UGV. This estimation will increase the accuracy of our map.

Secondly, we use the tensor voting framework to get the different features out of the cloud map. Thirdly, to get the exact position of the road curb, we use the DEM and feature filter to get points that belong to the road curb. In the fourth step we project these points to a 2D map and use different colors to represent the different sematics of the obstacles.

\subsection{Mapping method}

\begin{figure}[t!]
	\centering
	\includegraphics[width=0.45\textwidth]{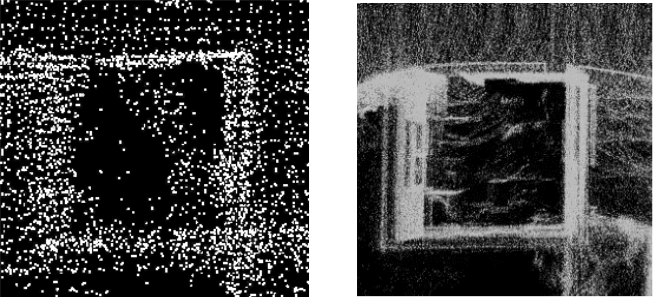}
	\caption{The comparison between LOAM and ICP.}
	\vspace{-1.5em}
\end{figure}

In the mapping process, we compare the difference be-tween LOAM and the ICP, as shown in Fig. 3. LOAM \cite{zhang2014loam} uses multi-beam LiDAR moving in 6-DOF and makes use of the line and plane features to match the different beams, while the ICP \cite{ye2017lidar} registers the point clouds by iteratively minimizing the distances between the closest points. The difference between the ICP and LOAM is the way of getting features: the ICP uses the geometric points information to match the features in different frames.

In Fig. 3 we compare the mapping method of LOAM (left) and the ICP mapping (right). We find that the ICP mapping has denser points compared to LOAM and it is more correct when it used for registration. The tensor voting framework, it will have better performance when we use a denser map.

\section{TENSOR VOTING FRAMEWORK}
\subsection{Tensor Voting Framework}
 Tensor voting was first used in image processing. This algorithm is based on the Gestalt theory, which means when people are looking at curves and surfaces, they will auto extend the curve and surface to an entirety.

Jia \cite{jia2003image} uses the feature of tensor voting for picture repairing, and translate the color and the texture information into the ND tensor to obtain the missing color. In recent research \cite{liu2012normal}, the tensor voting frame work is also used in the surface normal estimation for the point cloud obtained form LiDAR.

Normally, tensor voting has two tensor voting procedure, the first one is sparse voting, the second one is dense voting. Sparse voting only uses ball voting because at the beginning, each point does not have a tensor.

If the points are sufficiently dense and accurate enough, we can only use the sparse voting, because we do not need to reconstruct the surface.

\begin{equation}
k_{(d, \sigma)}=e^{-\frac{d^{2}}{\sigma^{2}}}
\label{a}
\end{equation}  

The sparse voting in the point cloud uses the location information as input to estimate the voting result. This procedure is called encoding, and gives each point 1 as the initial value. In general, this voting result obtains the refined tensors for another tensor voting process. The sparse voting is mainly calculating the ball voting. When we calculate the vote of other points, we choose to use Equation (1) to decay the influence of the points that are far away from current point, where d is the distance between the point and its neighbor, and $\sigma$ is the voting area.
\begin{equation}
\text { Ball }=\int_{0}^{2 \pi} \int_{0}^{2 \pi} R_{\gamma \alpha}^{-1}\left(T_{s} p\right) \mathrm{d} \theta_{1} \mathrm{d} \theta_{2}
\label{b}
\end{equation}  
 
Equation (2) shows the procedure to calculate ball voting.
$R_{\gamma \alpha}$ is a transform matrix, because we need to calculate the
other directions integration in the ball area. This is the result
of voting in different directions, and the point that we are
calculating. This formula represents that if a point is farther
away, the point that has been voted will receive less influence
from this point.
\begin{figure}[t!]
	\centering
	\includegraphics[width=0.5\textwidth]{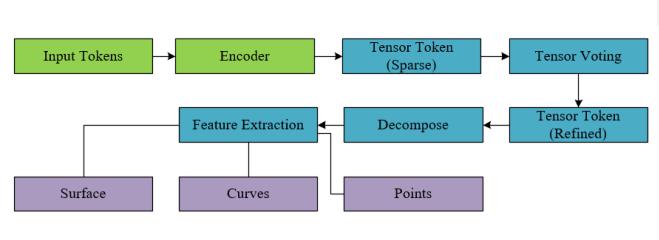}
	\caption{The steps of tensor voting.}
	\vspace{-1.5em}
\end{figure}

The algorithm runs as Fig. 4, which shows the procedure
of sparse tensor voting. We do not use dense voting because
of the time cost of the calculation.
 \begin{gather}
 T=\lambda_{1} \hat{e}_{1} \hat{e}_{1}^{\top}+\lambda_{2} \hat{e}_{2} \hat{e}_{2}^{\top}+\lambda_{3} \hat{e}_{3} \hat{e}_{3}^{\top} \\ \text { Stick component }=\left(\lambda_{1}-\lambda_{2}\right) \hat{e}_{1} \hat{e}_{1}^{\top} \\ \text { Plate component }=\left(\lambda_{2}-\lambda_{3}\right)\left(\hat{e}_{1} \hat{e}_{1}^{\top}+\lambda_{2} \hat{e}_{2} \hat{e}_{2}^{\top}\right) \\ \text { Ball component }=\lambda_{3}\left(\hat{e}_{1} \hat{e}_{1}^{\top}+\hat{e}_{2} \hat{e}_{2}^{\top}+\hat{e}_{3} \hat{e}_{3}^{\top}\right)
 \end{gather}

The tensor tokens we calculated after ball voting are in Equation (3), and the following step is to decompose these parts into a surface curves component. For the following component, we can use (4), (5) and (6) to decompose these features.

\begin{figure}[t!]
	\centering
	\includegraphics[width=0.40\textwidth]{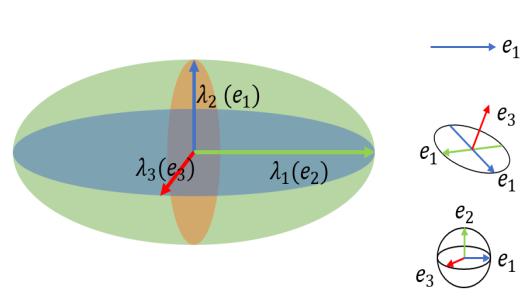}
	\caption{Tensor Voting component.}
	\vspace{-1.5em}
\end{figure}

Fig. 5 shows the tensor. Such a tensor can be visualized
as an ellipse in 2-D, or an ellipsoid in 3-D. The shape of the
tensor is ellipsoidal and it shows the captured information,
curve and surface element saliency.

In the tensor voting we regard $\lambda_{1} \geq \lambda_{2} \geq \lambda_{3}$  as the
eigen value. We also regard $\hat{e}_{1}\hat{e}_{2}\hat{e}_{3}$ as the eigen vector.

Fig. 5 shows the three components. The left component is
the stick tensor and it is in the shape of an elongated elli-
posid. We can regard it as a point on a smooth surface. The
middle component is the plate tensor, which is represented
by a circular disk. This disc is perpendiclar to the tangent
of the curve which is the junction of the two surfaces. The
right component is the ball tensor, which is an isolated point
or a junction of curves. This tensor does not have saliency
preference and the shape of the tensor is a sphere.

From Equation (4), (5) and (6) we can also get the different
saliencies; the stick votes are weighted by $\lambda_{1}-\lambda_{2}$  , the plate
votes by  $\lambda_{2}-\lambda_{3}$ and the ball votes by $\lambda_{3}$.

After decomposing the defeature, we give the additional
channels to the point. These channels stand for the saliency
of their different components. This step helps us to visulize
the result and get points which belong to the curb.

\subsection{Result of Tensor Voting}

\begin{figure}[t!]
	\centering
	\includegraphics[width=0.40\textwidth]{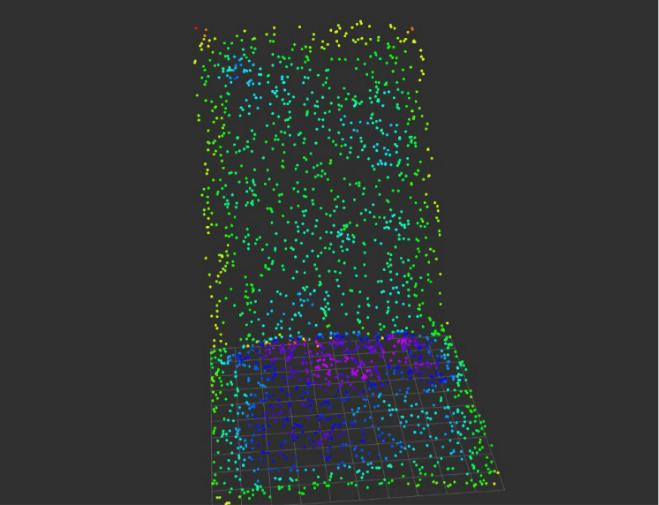}
	\caption{The test results of tensor voting.}
	\vspace{-0.5em}
\end{figure}

Fig. 6 shows the result of sparse tensor voting for a
test point cloud. This result shows the saliency of the stick
component. We can get the information that the bottom in
the colour purple shows the biggest possibility of belonging
to a surface.

We also tried the tensor voting algorithm on LOAM [14],
which can generate a dense map from an individual laser scan
from LiDAR. It calculates the transform between 2 frames and
adds each frame to be a big global map. We get the local
dense map to calculate the result.

\section{ROAD CURB DETECTION}

\begin{figure}[t!]
	\centering
	\includegraphics[width=0.40\textwidth]{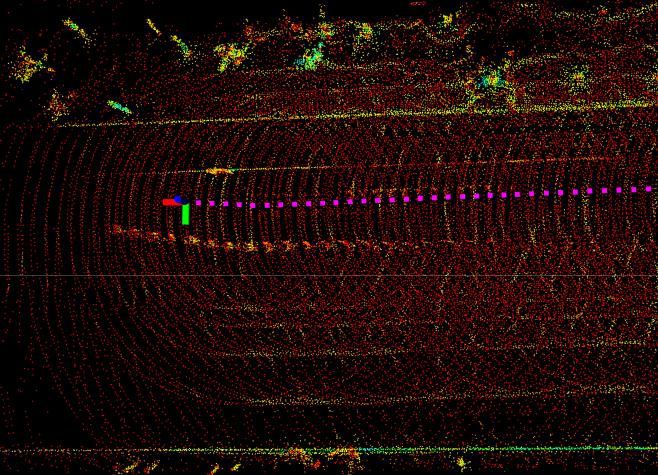}
	\caption{Tensor voting on local map.}
	\vspace{-1.5em}
\end{figure}

Fig. 7 shows the result after we cut the dense map
generated from LOAM into a $20 m \times 20 m \times 2 m$ cube
to test the result of tensor voting, we test it on the GTX
1080 GPU. We get an average calculation time of 0.34886 sec in a normal city situation. When we compare the LOAM mapping update speed, which is 1 HZ, it has the potential to be a real-time algorithm.

In this part we introduce the method to find the road using the result of tensor voting.

\subsection{Ground detection}
After we obtain the stick feature of the point cloud, we can extract the component which most likely belongs to the ground. We set a threshold of the stick component channel to get the plane and we can use the tensor voting result to estimate the grounds normal. \cite{liu2013information}

\begin{figure}[t!]
	\centering
	\includegraphics[width=0.40\textwidth]{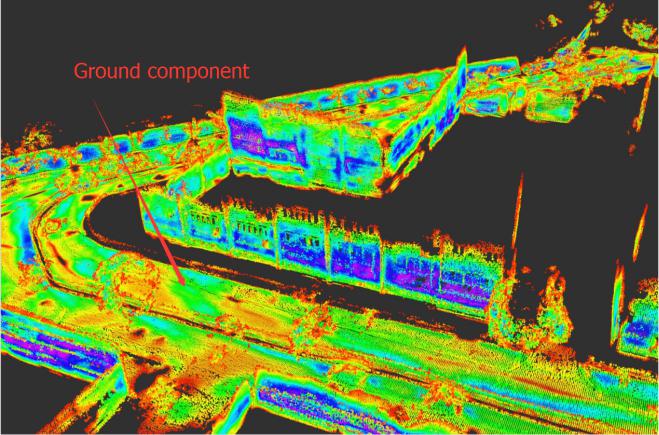}
	\caption{The result aftet surface extraction.}
	\vspace{-0.5em}
\end{figure}

Fig. 8 shows the result after we extract the ground com-ponent and other surface components. In further experiments it is important to know the exact height of the ground component as it helps to get the DEM.

\subsection{Line feature detection}
In the tensor voting framework it is easy to get the normal of the surface of each point by the direction of the feature value. After we get the surface normal, we can calculate the angle between the Z axis, which is perpendicular to the ground, and we can set a filter of the angle between each point normal and the Z axis.
Using this approach, we can obtain the initial estimate of the road height. We save these heights in the grid. We reduce the height that is different from the nearby grid as these noises come from surfaces like the top of trees, which also have same surface normal as ground.

The size of the grid should not be too large because if the surface is changing smoothly, and if the points are not dense enough it will lead to incorrect estimation of the height. In this environment, we choose the same size as the UGV which is $0.5 m \times 0.5 m$.

In the line feature detection, we can extract the line feature from the plate component. The line feature has a strong relation according to Equation (1), which describes how the area of other points can influence the current point.

In the tensor voting framework, it is easy to get the surface normal from the result, we can get the direction of surface. From the plate component (7) we find that the it shows the saliency $\lambda_{2}-\lambda_{3}$ and the surface direction.
\begin{equation}
T_{p}=\left(\lambda_{2}-\lambda_{3}\right)\left(\hat{e}_{1} \hat{e}_{1}^{\top}+\lambda_{2} \hat{e}_{2} \hat{e}_{2}^{\top}\right) 
\end{equation}

\begin{figure}[t!]
	\centering
	\includegraphics[width=0.40\textwidth]{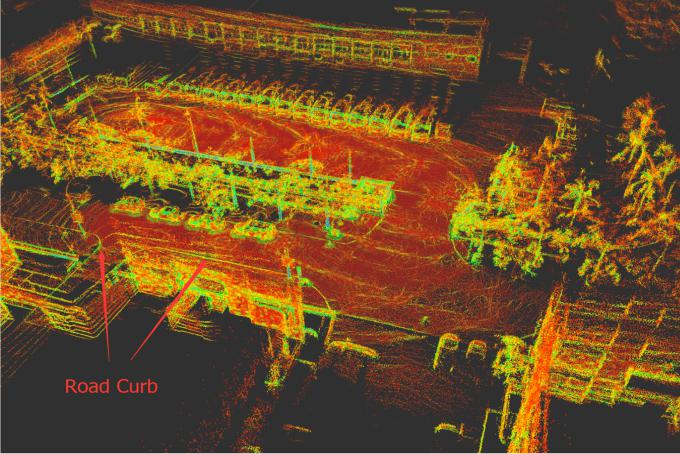}
	\caption{The line feature saliency.}
	\vspace{-1.5em}
\end{figure}

\begin{figure}[t!]
	\centering
	\includegraphics[width=0.40\textwidth]{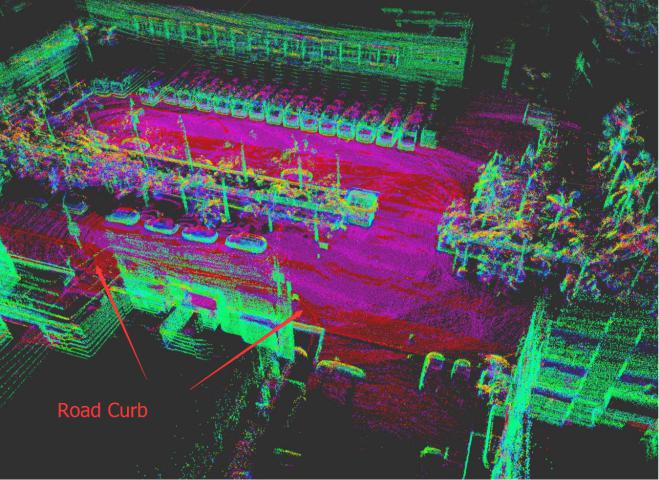}
	\caption{The Z axis saliency of each points.}
	\vspace{-0.5em}
\end{figure}

As is shown in Fig. 9, we can obtain the road curb, which is green. The green part in the point cloud are points with a higher $\lambda_{2}-\lambda_{3}$ value; the higher value means that point is more likely to be at a line. Fig. 10 shows the Z axis volume, this is the result that is directly obtained from the tensor voting algorithm. It can be regarded as the saliency of the vertical and we can find the road curb has different vertical intensities compared to other points with a low plate intensity. This value also can be regarded as the z-component normal of the point cloud.

\subsection{Filtering}

After we detect the line feature and the surface normal, we use the DEM to be the height estimation of the ground. The DEM is used to describe the ground height of the map.

In the first step, we get a rough estimation of the ground which is based on the height of the plane and surface direction. The rough estimation helps us to reduce large errors of the ground. We set it as $10 m \times 10 m$, then we can get the accurate estimation of the elevation of the ground by reducing the grid which may have a large error with the majority ground components. We set it as $1 m \times 1 m$. From this approach we can get a good height estimate of each point.

We use the refined elevation map to reduce the unrelated points, such as the upper part of trees and some line features of other vehicles. We regard the height that is 50 cm higher than the ground as unlikely to find the curb.
\begin{figure}[t!]
	\centering
	\includegraphics[width=0.40\textwidth]{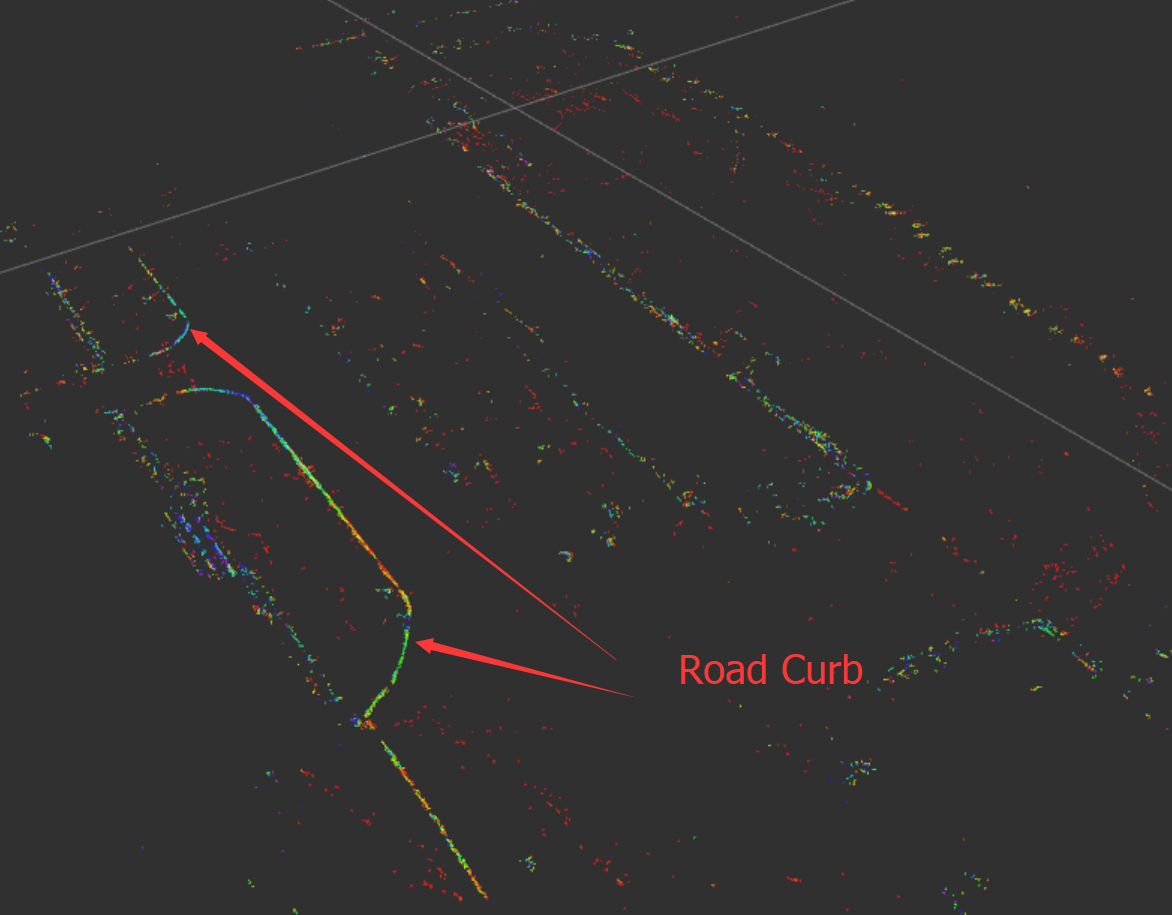}
	\caption{The points that belong to road curb.}
	\vspace{-1.5em}
\end{figure}

\begin{figure}[t!]
	\centering
	\includegraphics[width=0.40\textwidth]{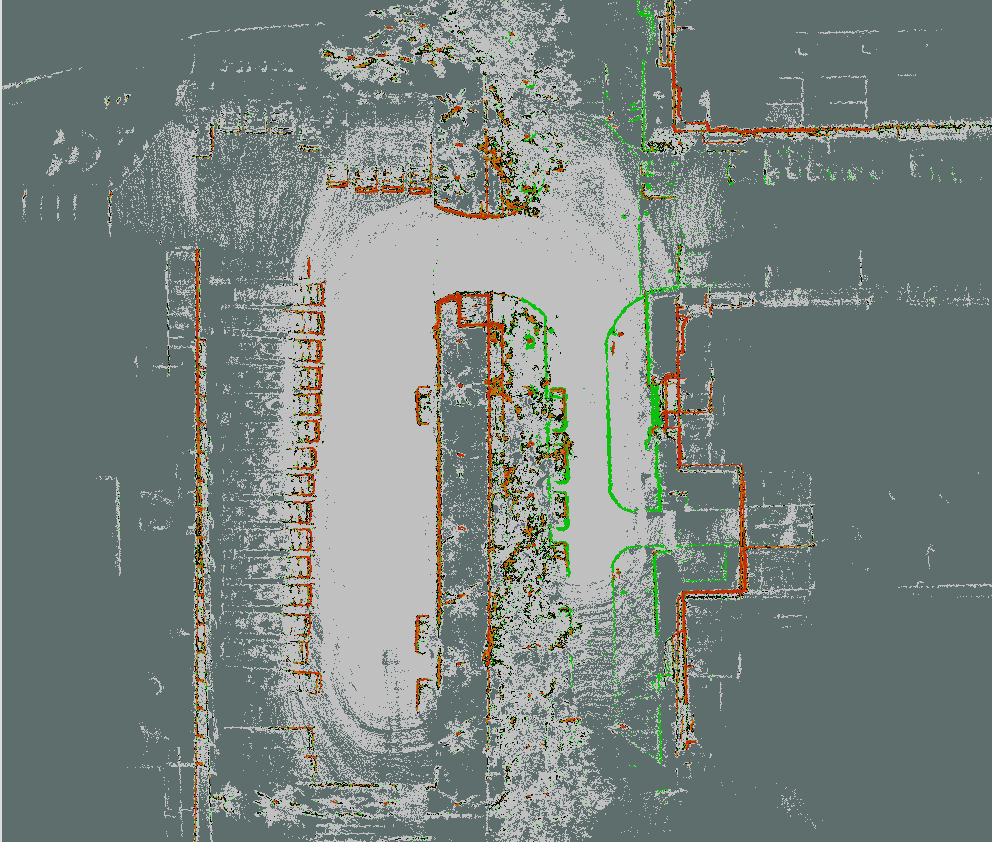}
	\caption{The experimental results. The point clouds in red are
the detected walls. The point clouds in green are the detected
road curbs.}
	\vspace{-1.5em}
\end{figure}

The result is shown in Fig. 11. After we use the line feature and the DEM to reduce the unrelated area, we can find that the majority of the road curb has been extracted successfully. Then we use the outlier filter to strengthen the road curb.
\subsection{Project to map}
The 2D grid map has good potential to formulate a high definition map. This kind of map has taken up a small amount of storage memory and the grid map can has a good sematic when giving different colors.

In this part, we use the result of tensor voting, the number of points in same grid, and the road curb to get the different sematics of the map. In the Fig. 12 we can see the road curb and other obstacles with a small height, which are also marked in green in the 2D grid map. An obstacle that is absolutely not traversable, such as a wall or vehicle, is marked in red and the zone sufficient enough points and tensor voting results we regard as the unknown area. The details judge feature and the different sematics are shown in the TABLE. 1.

\begin{table}[htbp]
\caption{Different Feature And Sematic}
\begin{center}
\begin{tabular}{|c|p{3cm}|c|p{1.5cm}|}

\hline
\textbf{Sematic}&\multicolumn{3}{|c|}{\textbf{Feature and color}} \\
\cline{2-4} 
\textbf{Head} & \textbf{\textit{Feature}}& \textbf{\textit{Color}}& \textbf{\textit{Traversable}} \\
\hline
Road curb& Result of Tensor Voting & Green& Certain conditions \\
\hline
Obstacle&Not higher than UGV&Black& No\\
\hline
Wall/Vehicle& Obstacle has lots of points in high position& Red & No\\
\hline
Road&Surface normal has same direction with Z axiz andclosed to the hight of DEM&Gray &Yes\\
\hline
Unknown&Without enough points to do tensor voting&Dark green&Yes\\
\hline
 
\end{tabular}
\label{tab1}
\end{center}
\end{table}

We also make a definition about the trafficability of each sematic, this helps the path planning for further study. We can cross the road curb with a lower speed and with a lower angle, this will help us to have more options when we want to go from point A to B.
The sematic map will also help us to achieve localization because it has more different features than a normal grid map, and it also takes a lower CPU utilized percentage which means it is easier for global path planning.
\section{CONCLUSION}
In this paper, we aim to solve the problem of obtaining the road curb from the data of a LiDAR sensor. This problem is critical in formulating a high definition map and for urban autonomous driving. We also built a 2D sematic map to make use of the different features.

We first generate the dense map from the ICP mapping method and we compare two different mapping methods. Then we present a new method based on tensor voting to extract the line feature and surface feature after mapping. Next we use the digital elevation map to get the basic estimation of the average height of each grid as the further parameters of the filter. We also set the different feature filters, to get the correct points of the road curb.

Additionally, we focus on how to get a better sematic of the map in future usage. In this approach, we refine the map to a resolution of 0.12 meters, which meets our expectation. We also find that the storage of the 2D grid mapping is just 707 kB, this result is 85.4 times smaller than the result of ICP mapping and 9.7 times smaller than the result of LOAM.

In the future, we will add some details like lane lines into this sematic map to get more precise and abundant information about the road condition. With more information, we can obtain more rules to control vehicles in a high speed environment in the future.

Additionally, another point we should like to improve in the future is that we should set some regions of interest (ROI) for our further tensor voting process. This approach will reduce the calculation complexity and finally realize real-time road curb detection.
\section*{Acknowledgment}
 This work is supported by grants from the Research Grants Council of the Hong Kong SAR Government, China (No. 11210017, No. 16212815, No. 21202816, NSFC U1713211) awarded to Prof. Ming Liu. This work is also supported by grants from the Shenzhen Science, Technology and In-novation Commission, JCYJ20170818153518789, and Na-tional Natural Science Foundation of China (No. 61603376) awarded to Dr. Lujia Wang.
\bibliographystyle{IEEEtran}
\balance

\end{document}